\documentclass{article}
\usepackage{spconf,amsmath,graphicx}
\usepackage{booktabs}
\usepackage{caption}
\usepackage{subcaption}
\usepackage{placeins}
\usepackage{tikz}
\usepackage{setspace}
\usepackage{hyperref}
\usepackage{cleveref}

\usetikzlibrary{calc}
\usetikzlibrary{decorations.pathreplacing}

\definecolor{uibkblue}{cmyk}{1,0.6,0,0.65}%
\definecolor{uibkorange}{cmyk}{0,.5,1,0}%
\definecolor{uibkgray}{cmyk}{0,0,0,0.6}%
\definecolor{red}{rgb}{0.8,0.06,0.00}%

\tikzstyle{color1} = [color=red]
\tikzstyle{color2} = [color=uibkorange]
\tikzstyle{color3} = [color=uibkblue]

\DeclareMathOperator{\sign}{sign}

\providecommand{\cameraready}[1]{}

\newcommand*\rot{\rotatebox{90}}

\title{Forensicability of Deep Neural Network Inference Pipelines}
\name{Alexander Schlögl \qquad Tobias Kupek \qquad Rainer Böhme}
\address{University of Innsbruck, Austria}

\begin{document}
%
\maketitle
\begin{abstract}
  We propose methods to infer properties of the execution environment of machine learning pipelines by tracing characteristic numerical deviations in observable outputs.
  Results from a series of proof-of-concept experiments obtained on local and cloud-hosted machines give rise to possible forensic applications, such as the identification of the hardware platform used to produce deep neural network predictions.
  Finally, we introduce boundary samples that amplify the numerical deviations in order to distinguish machines by their predicted label only.
\end{abstract}

\begin{table}[b!]
  \begin{singlespace}
    \par\noindent\rule{\linewidth}{0.4pt}
    \footnotesize

    Copyright 2021 IEEE.
    Published in ICASSP 2021 - 2021 IEEE International Conference on Acoustics, Speech and Signal Processing (ICASSP), scheduled for 6-11 June 2021 in Toronto, Ontario, Canada.
    Personal use of this material is permitted.
    However, permission to reprint/republish this material for advertising or promotional purposes or for creating new collective works for resale or redistribution to servers or lists, or to reuse any copyrighted component of this work in other works, must be obtained from the IEEE.
    Contact: Manager, Copyrights and Permissions / IEEE Service Center / 445 Hoes Lane / P.O. Box 1331 / Piscataway, NJ 08855-1331, USA. Telephone: + Intl. 908-562-3966.
  \end{singlespace}
\end{table}
%
\section{Forensics of Deep Learning Inference}
\label{sec:intro}
Over the past 20 years, media forensics has matured as a sub-field of signal processing specialized on the extraction of forensically relevant information from digital signals~\cite{Chu2016Limits,Chu2016Detectability}.
Relevant applications include the proof of authenticity (or forgery) of image, audio, or video data.
The past decade saw a surge in the use of machine learning (ML), including deep neural networks (DNNs), as a \emph{method} in media forensics~\cite{Yang2020Survey}.
Here we change the perspective and ask if machine learning pipelines themselves can be \emph{subject} to forensic analysis.

Recall that media forensics relies on the statistical detection of traces introduced by characteristic quantization noise of a sequence of signal processing operators, for instance to identify a specific digital camera as the source of an image under analysis~\cite{BK2016-IHBOOK}.
Likewise, modern DNN inference pipelines are concatenated signal processing operators that apply quantization to activation values at every layer of a deep network.
Results of our \cameraready{preliminary }experiments show that differences in the implementation of the inference pipelines indeed leave characteristic traces in the network's response, which allow a forensic investigator to detect the specific software and hardware configuration used to evaluate a given DNN model.

This observation has several practical implications.
First, the method could reveal the execution environment used for machine classification, thereby improving transparency and allowing a subject of an automated decision to verify the context in which the decision was made~\cite{Tram2019Slalom}.
Second, customers of machine learning-as-a-service could test if they are provided with the technical platform they have rented~\cite{Ghodsi2017Safetynets}.
Third, outputs of generative models, such as the controversial DeepFakes~\cite{Korshunov2018Deepfakes}, could be traced back to the creator; or at least the search space could be reduced by ruling out unlikely origins.

The paper proceeds as follows.
Section~\ref{sec:framework} presents a framework defining the sources of characteristic traces in a typical DNN inference pipeline.
We provide early validation by instantiating the framework with a series of experiments.
Three common image classification networks are run in nine execution environments in order to find out if it is possible to:
\begin{enumerate}
  \item identify the execution environment of the complete classification pipeline from its real-valued outputs, e.\,g., confidence vectors (Section~\ref{ssec:complete});\\[-4ex]
  \item attribute forensic traces to the execution plan vis-a-vis the actual arithmetic units (Section~\ref{ssec:plan});\\[-4ex]
  \item explain the generation of forensic traces with properties of the model and the input data (Section~\ref{ssec:complexity});\\[-4ex]
  \item leverage techniques inspired by adversarial samples (inputs close to or just across the decision boundary~\cite{Szegedy2014Intriguing}) to identify classification pipelines from categorial outputs, i.\,e., class labels (Section~\ref{ssec:ae}).
\end{enumerate}
We find affirmative answers to all four questions.
The final Section~\ref{sec:discussion} concludes with a discussion and outlook.

\section{Framework}
\label{sec:framework}

\begin{figure*}
  \resizebox{\textwidth}{!}{%
    \tikzstyle{phase} = [rectangle, thick, text centered, draw=uibkgray]
\tikzstyle{process} = [rectangle, thick, minimum width=2.25cm, minimum height=1.25cm, text centered, draw=black]
\tikzstyle{part} = [circle, thick, minimum width=0.75cm, minimum height=0.75cm, text centered, draw=black]
\tikzstyle{arrow} = [thick,->,>=latex]

\begin{tikzpicture}
    \node[color=uibkgray] at (1, -1.3) {training phase};
    \draw[phase] (6.55, -1.6) rectangle (17.5,2);
    \draw[phase, dotted] (17.5, -1.6) -- (22.3,-1.6) -- (22.3,2) -- (17.5,2);
    \node[color=uibkgray] at (8, -1.3) {inference pipeline};
    
    \node[] (start) {};
    \node[process, right of=start, xshift=3cm] (train) {train};
    \node[process, right of=train, xshift=4cm] (prepare) {prepare};
    \node[process, right of=prepare, xshift=4cm] (compute) {compute};
    \node[process, right of=compute, xshift=4cm] (argmax) {argmax};
    \node[right of=argmax, xshift=3cm] (end) {};
    
    \draw [arrow] ($(start)+(0,-2mm)$) -- node[above, yshift=4mm] {model architecture\strut} ($(train.west)+(0,-2mm)$);
    \draw [arrow] ($(start)+(0,2mm)$) -- node[below, yshift=-4mm] {training data} ($(train.west)+(0,2mm)$);
    \draw [arrow] (train) -- node[above] {trained model} (prepare);
    \draw [arrow] (prepare) -- node (mid) {} node[above] (compplain) {compute plan} (compute);
    \draw [arrow] (compute) -- node (mid2) {} node[above, text width=2cm, align=center] {real-valued outputs} (argmax);
    \draw [arrow] (argmax) -- node (mid3) {}node[above] {label} (end);
    
    \node[above of=train, xshift = -1cm, yshift=1.5cm, color1] (HW1) {HW};
    \node[above of=train, yshift=1.5cm, color2] (SW1) {SW};
    \node[above of=train, xshift = 1cm, yshift=1.5cm, color3, align=center] (seed) {random\\ seed};
    \draw [arrow, color1] (HW1) |- ($(train.north)+(-2mm,0.6cm)$) -- ($(train.north)+(-2mm,0)$);
    \draw [arrow, color2] (SW1) -- (train.north);
    \draw [arrow, color3] (seed) |- ($(train.north)+(2mm,0.6cm)$)-- ($(train.north)+(2mm,0)$);
    
    \node[above of=prepare, yshift=0.6cm, color3] (inputprop) {input properties};
    \node[above of=mid, yshift=1.5cm, color1] (HW2) {HW};
    \node[above of=mid, xshift = 1cm, yshift=1.5cm, color2] (SW2) {SW};
    \node[below of=compute, yshift=-1.1cm,] (input) {input};
    \draw [arrow, color3] (inputprop) -- (prepare);
    \draw [arrow, color1] (HW2) |- ($(prepare.north)+(2mm,0.6cm)$)-- ($(prepare.north)+(2mm,0)$);
    \draw [arrow, color2] (SW2) |- ($(prepare.north)+(4mm,0.4cm)$)-- ($(prepare.north)+(4mm,0)$);
    \draw [arrow, color1] (HW2) |- ($(compute.north)+(-2mm,0.6cm)$)-- ($(compute.north)+(-2mm,0)$);
    \draw [arrow, color2] (SW2) |- ($(compute.north)+(-4mm,0.4cm)$)-- ($(compute.north)+(-4mm,0)$);
    \draw [arrow] (input) -- (compute);
    
    \node[part, below of=mid] (A) {A};
    \node[part, below of=mid2] (B) {B};
    \node[part, below of=mid3] (C) {C};

\end{tikzpicture}
  }
  \caption{System model for the forensic analysis of DNN inference pipelines. Arrows from the top indicate causes of characteristic traces. The main experiments compare equivalence classes of the real-valued outputs (B) for given trained models and input.
    One experiment decomposes the pipeline by looking for traces in the compute plan (A).
    Another experiment takes control of the input in order to propagate forensically useful information to the label (C). The training phase is shown for completeness.}
  \label{fig:framework}
\end{figure*}
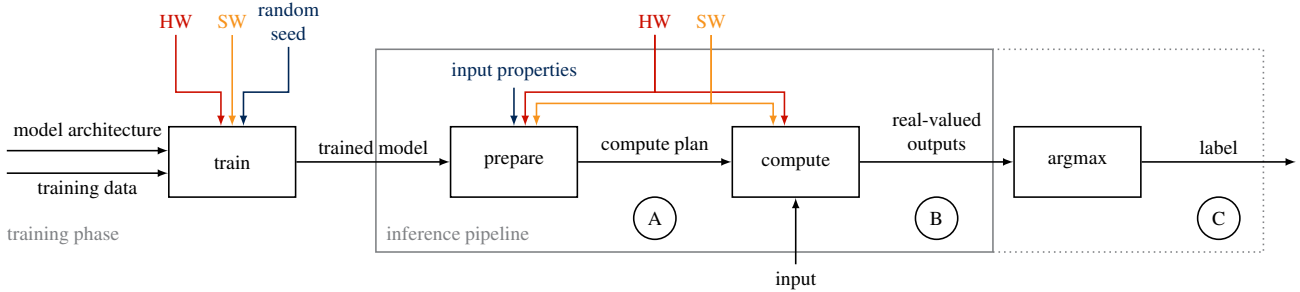

Figure~\ref{fig:framework} shows our system model for analyzing ML pipelines from a signal processing perspective.
It broadly separates training and inference phase, and intended parameters from unintended factors influencing the output.
These factors include the hardware (HW) and software (SW) of the execution environment.
Although most platforms follow the IEEE-754 standard for floating-point arithmetics~\cite{kahan1996ieee}, outputs may differ between execution environments.
Causes include arithmetic units using faster approximations of IEEE-754 (like in some GPUs), intermediate values being rounded to fit in registers of different precision (e.\,g., for SIMD instructions), or transformations made during execution planning.
Tuning tricks like loop unrolling, constant folding, and arithmetic simplifications are common in performance-optimized ML toolboxes and may also vary between versions.


The present work focuses on the inference pipeline and assumes that the trained model is known to the forensic analyst.
It consists of the model architecture and weights.
The analyst's task is to distinguish different execution environments by observing the outputs of the inference pipeline for a given (known) input.
Real-valued outputs observable at point (B), such as logit or confidence vectors, carry more forensically relevant information.
This is so because small numerical deviations likely disappear after the argmax operator, which discretizes the result to a class label (C).

While the analyst, in general, cannot observe intermediate values or any other side-channel (e.\,g., timing) from the inference pipeline, we also define point (A) to study the influence of the compute plan as a cause of traces.
The plan in turn may depend on input properties, such as the number and dimensions of input objects.
Many ML toolboxes let developers inspect the compute plan.

\section{Experiments}
\label{sec:experiments}


We explore selected factors experimentally on a set of different machines (we will refer to them as Architectures, \emph{Arch}), as listed in \Cref{tbl:hosts}.
Architectures i to iii are physical instances on our premises, while iv to ix are cloud instances of the Google Cloud Platform (GCP) and Amazon Web Services (AWS), respectively.
While we know exactly what CPUs are installed on our machines, the cloud providers only let us specify the CPU family and a lower bound for the generation.

\begin{table}[h!]
  \centering
  \caption{List of tested CPU architectures} 
  \begin{tabular}{@{}c@{~~}l@{~~}l@{~~}l@{~}lr@{}}
    \toprule
    \textbf{Arch} & \textbf{Type} & \textbf{Vendor} & \textbf{Generation} & \multicolumn{2}{@{}p{35mm}@{}}{\textbf{CPU} \hfill \textbf{Memory (GB)}} \\
    \midrule
    i             & local         & Intel           & Coffee Lake         & i7-9700      & 32       \\
    ii            & local         & Intel           & Kaby Lake           & Xeon E3-1270 & 32       \\
    iii           & local         & AMD             & Ryzen               & TR 2950X     & 126      \\
    iv            & GCP           & Intel           & Broadwell           & Xeon E5 vCPU & 3.75     \\
    v             & GCP           & Intel           & Sandy Bridge        & Xeon E5 vCPU & 3.75     \\
    vi            & GCP           & Intel           & Skylake             & Xeon vCPU    & 3.75     \\
    vii           & AWS           & Intel           & Broadwell           & Xeon E5 vCPU & 4        \\
    viii          & AWS           & Intel           & Skylake             & P-8175 vCPU  & 4        \\
    ix            & AWS           & Intel           & \emph{N.A.}         & Xeon vCPU    & 4        \\ \bottomrule
\end{tabular}
  \label{tbl:hosts}

\end{table}

We use one fixed toolbox, popular TensorFlow, at version 2.3.0.
We tested other versions as well but found every version since 1.1.15 returning equivalent results.
All results are consistent over multiple executions.
We use three standard datasets for our experiments: MNIST handwritten digit recognition~\cite{LeCun2010MNIST}, Fashion-MNIST (FMNIST) object recognition~\cite{Xiao2017FMNIST}, and CIFAR-10 object recognition~\cite{Krizhevsky2009CIFAR}.
One model architecture is trained per dataset.
We use a small CNN with two convolutional and one dense layer for MNIST, a 20-layer deep variant of ResNet~\cite{He2016ResNet} for FMNIST, and a 32-layer deep variant for CIFAR-10.
We train one model per dataset and use it for all experiments in this study.


\subsection{Identification of Complete Inference Pipelines}
\label{ssec:complete}
We perform inference on a test object for each model on all architectures in the study.
The resulting outputs are separated into equivalence classes, as shown in~\Cref{tbl:overview}.
Prediction outputs within the same equivalence class are \emph{binary equivalent}.
Letters referring to equivalence classes are distinct between models (columns) but should not be compared across tables.
The results show that even this small sample of architectures falls into at least four classes in such a way that processor generations can be identified uniquely.

\begin{table}[h!]
  \centering
  \caption{Main result: equivalence classes of architectures}
  \begin{tabular}{l r  r  r   c} 
        \toprule
        \textbf{Dataset}            &
        \multicolumn{1}{c}{MNIST}   &
        \multicolumn{1}{c}{FMNIST}  &
        \multicolumn{1}{c}{CIFAR-10} &
        \\

        \midrule

        Arch i                      & A & E & K & \\
        Arch ii                     & A & E & K & \\
        Arch iii                    & B & F & L & \\
        Arch iv                     & B & G & M & \\
        Arch v                      & C & H & N & \\
        Arch vi                     & D & I & O & \\
        Arch vii                    & B & J & P & \\
        Arch viii                   & D & I & O & \\
        Arch ix                     & B & G & M & \\
        \bottomrule
\end{tabular}
  \label{tbl:overview}
\end{table}


Our two locally hosted architectures i and ii produce identical results (recall that Coffee Lake is a minor refinement of the Kaby Lake architecture), as do the two Skylake processors vi and viii.
We find identical results for iv and ix, where the generation for ix is not specified.
The two Broadwell processors iv and vii produce different results, with vii forming its own equivalence class.
Remember that we cannot rule out that the cloud provider assigned instance vii a higher processor generation than Broadwell.
Unsurprisingly, the AMD processor iii is distinguishable from all Intel processors for the more complex models, as is the single Sandy-Bridge processor v.
Our takeaway is that passive forensics of deep neural network inference pipelines is possible in principle.

\subsection{Effect of the Execution Plan}
\label{ssec:plan}
In this section, we decompose the inference pipeline into two stages to study the influence of the preparation of the compute plan from the actual computation.
TensorFlow Lite (TFLite) allows us to run the preparation step on its own and store the generated compute plan.
TFLite is intended for mobile and embedded platforms that are too constrained to optimize the compute graph on-the-fly.
We generate TFLite compute plans on our three local architectures and verify that the process is deterministic.
Each compute plan is then executed on all local architectures, resulting in nine predictions in total.

Executing a single compute plan on different architectures yields identical results in all tested cases.
However, compute plans generated on different architectures produce different results \emph{even when executed on a single architecture}.
\Cref{tbl:tflite}~shows the resulting equivalence classes.

\begin{table}[h]
  \centering
  \caption{Effect of the execution plan (on local architectures)}
  \begin{tabular}{l  @{~}c @{~}c @{~}c @{}r}
    \toprule
    \textbf{Dataset}            &
    \multicolumn{1}{c}{MNIST}   &
    \multicolumn{1}{c}{FMNIST}  &
    \multicolumn{1}{c}{CIFAR-10} &
    \\

    \midrule

    Arch i                      & A & C & F & \\
    Arch ii                     & A & D & G & \\
    Arch iii                    & B & E & H & \\
    \bottomrule
\end{tabular}
  \label{tbl:tflite}
\end{table}

The simple MNIST model produces identical results for architectures i and ii, running Intel processors.
Generating the plan on an AMD processor leads to different results.
For our more complex models, all architectures return different results, unlike in~\Cref{tbl:overview}.
TFLite compute plan generation also optimizes the model in some ways, folding constants and even slightly pruning the model.
These optimizations seem to be more sensitive to hardware differences than regular TensorFlow inference.
Upon closer inspection we also find that the equivalence classes of TFLite compute plans are isomorphic to the classes of their predictions.

\subsection{Effect of Model and Input Properties}
\label{ssec:complexity}

The results in \Cref{ssec:complete} have revealed that some model architectures leave more traces than others. For example in \Cref{tbl:overview}, Arch~iii and iv fall into different equivalence classes for the FMNIST and CIFAR model, but not for MNIST.
This leads us to explore the effect of model complexity.

We first establish that traces emerge chiefly in the computation of convolutions, as we do not observe any differences in the predictions of \emph{all} tested architectures if we reduce the MNIST model to a simple multilayer perceptron (MLP) without convolutional layers.
%
We then create a set of simple mock models of varying complexity, and compare the resulting equivalence classes.
These models are initialized with weights depending on a fixed seed, but not trained, as we do not care about prediction accuracy.
We infer the models on a single MNIST sample input. Table~\ref{tbl:complexity} shows the equivalence classes for the reference models, a single convolutional layer, and four models with two consecutive convolutional layers each. The single convolution layer consists of 32 filters with a $3 \times 3$ kernel. The other four models vary in the number of filters ($1$ or $2$) and the kernel size ($2 \times 2$, $3 \times 3$ or $4 \times 4$). No activation function is used for the respective layers.

\begin{table}[h]
  \centering
  \caption{Impact of model complexity on the forensicability of MNIST inference pipelines. The mock models vary in complexity, but none is as complete as the reference CNN.}
  \label{tbl:complexity}
  \begin{tabular}{c c  c  c  c  c  c  c   c} 
    \toprule
    \multicolumn{9}{r}{\tikz[>=stealth]{\draw [->] (0,0)-- node [above] {\small increasing model complexity} ++(5cm,0);}\qquad~}
    \\
    \textbf{Arch}                                       &
    \multicolumn{1}{c}{\rot{Ref.~CNN}}                  &
    \multicolumn{1}{c}{\rot{MLP}}                       &
    \multicolumn{1}{c}{\rot{1 Conv}}                    &
    \multicolumn{1}{c}{\rot{\begin{tabular}[c]{@{}l@{}}2 Conv\\$1 \times 2 \times 2$\end{tabular}}} &
    \multicolumn{1}{c}{\rot{\begin{tabular}[c]{@{}l@{}}2 Conv\\$1 \times 3 \times 3$\end{tabular}}} &
    \multicolumn{1}{c}{\rot{\begin{tabular}[c]{@{}l@{}}2 Conv\\$1 \times 4 \times 4$\end{tabular}}} &
    \multicolumn{1}{c}{\rot{\begin{tabular}[c]{@{}l@{}}2 Conv\\$2 \times 4 \times 4$\end{tabular}}} &                             \\
    \midrule

    i & A & E & F & G & I & L & N &\\
    ii & A & E & F & G & I & L & N &\\
    iii & B & E & F & G & I & L & N &\\
    iv & B & E & F & G & I & L & N &\\
    v & C & E & F & G & K & M & P &\\
    vi & D & E & F & H & J & L & O &\\
    vii & B & E & F & G & I & L & N &\\
    viii & D & E & F & H & J & L & O &\\
    ix & B & E & F & G & I & L & N &\\

    \bottomrule
\end{tabular}

\end{table}

We observe that the single layer is not sufficient to cause a disparate output. When using two convolutional layers, already a single $2 \times 2$ filter leads to different outputs on architecture vi and viii. Increasing the kernel size to $3 \times 3$ adds an additional equivalence class for v. Surprisingly, when increasing the kernel size to $4 \times 4$, this extra class is not present. Using the same $4 \times 4$ kernel size for two instead of one convolution filter, the outputs show all three classes.
These experiments indicate inference pipelines become more distinguishable if predictions are done on more complex models, in terms of the number and type of convolutions.

Convolutional layers extract spatial features by sliding a kernel over the input, followed by element-wise multiplication and summartion~\cite{LeCun1998Gradient}. This operation depends on input properties, such as the dimensions, which might influence the resulting equivalence classes. We could observe this in preliminary tests, but leave a systematic analysis for future work.

%
%
%

\subsection{Amplifying Forensic Information}
\label{ssec:ae}

The fourth experiment turns to the effect of the input data.  
We investigate if a forensic investigator can craft specific samples that amplify the forensically relevant information such that the sought characteristic is propagated from the real-valued confidence vectors to the class label. Drawing on methods for generating adversarial samples, we generate samples as close as possible to the decision boundary.
We will refer to these samples as \emph{boundary samples}.
To generate boundary samples, we use an iterative version of the fast gradient sign method (FGSM), initially developed by Goodfellow et al.~\cite{goodfellow2014Harnessing}, and improved by Kurakin et al.~\cite{kurakin2017Physical}.
Iterative FGSM allows us to start from a test sample and approach the decision boundary by repeatedly applying untargeted adversarial perturbations.
The individual perturbation steps can be calculated as,
\begin{equation}
  \mathbf{x}_i = \mathbf{x}_{i-1} + \alpha\ \sign(\nabla_{\mathbf{x}} m(\mathbf{x}_i)),
  \label{eq:fgsmstep}
\end{equation}
where $\mathbf{x}_{\text{in}}$ is the input sample, $\mathbf{x}_i$ the adversarial sample generated in step $i$, $m$ is the model, and $\alpha$ is a chosen step size.

We modify \Cref{eq:fgsmstep} in two ways.
The step size $\alpha$ has an additional coefficient $\delta_{\text{conf}}$ calculated as the difference between the highest and second-highest confidence.
We also add a second coefficient to the gradient sign $l \in \{-1,1\}$, depending on the correctness of the classification of $\mathbf{x}_i$.
If $\mathbf{x}_i$ is correctly classified, we apply the normal adversarial perturbation.
If it is incorrectly classified this means we have crossed the decision boundary.
We then switch the sign of the gradient and move back towards the original class, similar to Newton approximation.
This process is repeated until the difference in confidences is sufficiently small ($10^{-8}$ in our experiments), or the maximum number of iterations is reached ($300$).

Using the modified FGSM step,
\begin{equation}
  \mathbf{x}_i = \mathbf{x}_{i-1} + l\ \alpha\ \delta_{\text{conf}} \sign(\nabla_{\mathbf{x}} m(\mathbf{x}_i)),
  \label{eq:advgeneration}
\end{equation}

we first generate boundary samples for one randomly drawn image from the test set of each dataset.
These boundary samples have a have a close-enough highest and second-highest confidence to cause label flips between architectures.

\Cref{fig:boundary} reports the confidences, labels, and confidence differences.
Observe that with control over the inputs it is possible to amplify prediction differences such that they are observable even with label outputs only.
The PSNR values are 25.55\,dB for MNIST, 37.69\,dB for FMNIST, and 57.18\,dB for CIFAR-10.
We also generate batches of 100 boundary samples for each dataset, achieving label flips in 48\,\%, 12\,\%, and 3\,\% for MNIST, FMNIST, and CIFAR-10, respectively.
This indicates that finding boundary samples is practical and, according to their PSNR, they could pass a manual inspection for plausibility.

\begin{table}[h]
  \centering
  \caption{Predictions for the proposed boundary samples.}
  \label{fig:boundary}
   \begin{tabular}{c@{~}c@{~}r@{~}r@{~}r}
	\toprule
    & \textbf{Label} & \textbf{Confidence} & \textbf{2\textsuperscript{nd} Confidence}   & \textbf{Difference}    \\
    \midrule
    \multicolumn{5}{l}{\textbf{MNIST}}\\
    Arch i       & 9     & \small 0.49980018 & \small 0.49979922 & \small 9.53e-07      \\
    Arch iii     & 4     & \small 0.49979982 & \small 0.49979958 & \small 2.38e-07 \\
    \midrule
    \multicolumn{5}{l}{\textbf{FMNIST}}\\
    Arch i       & 4     & \small 0.31779698 & \small 0.31779516 & \small 1.81e-06   \\
    Arch iii     & 0     & \small 0.31779578 & \small 0.31779578 & \small 0.0        \\
    \midrule
    \multicolumn{5}{l}{\textbf{CIFAR10}}\\
    Arch i       & 3     & \small 0.29437256 & \small 0.2943725 & \small 5.96e-08   \\
    Arch iii     & 5     & \small 0.2943726  & \small 0.2943721 & \small 4.76e-07   \\
    \bottomrule
    \multicolumn{5}{l}{\small Arch~ii is omitted because it is identical to Arch~i, as in \Cref{tbl:overview}}
\end{tabular}
\end{table}

\section{Discussion and Outlook}
\label{sec:discussion}

Due to space constraints, this conference paper is nothing more than a peek into a novel perspective for signal forensics.
Instead of using DNNs merely to extract forensic information from conventional processing pipelines, we make the DNN's inference pipeline the subject of forensic investigations.
Our results, obtained in a straight-forward manner without cherry-picking favorable settings,
\footnote{Code, models, data, and boundary samples are available at \url{https://github.com/alxshine/foreNNsic}}
demonstrate the general feasibility of the approach and leave a heap of new questions for future investigations.


The most obvious direction of future work is to scale up the analysis.
GPUs are promising targets because they are frequently used for inference.
Embedded devices, ranging from small IoT devices to smartphones and tablets, are also becoming ML platforms nowadays.
More fundamental new directions include
an extension to generative models~\cite{Goodfellow2013MPDBM,Karras2019stylegan2},
an exploration of blind variants that do not require knowledge of the trained model,
and an information-theoretical analysis of the forensicability, along the lines taken by Chu et al.~\cite{Chu2016Limits} or Pasquini et al.~\cite{pasquini19Bounds} for ``conventional'' media forensics.

Our method to create boundary sample can be optimized in at least two directions.
One is to increase forensic information by activating more neurons in the model~\cite{He2019SensitiveFingerprinting}.
The other is to evade detection methods put in place against adversarial samples~\cite{Li2017Detection,Xu2018Squeezing,SSPB2018-EUSIPCO}.
The fact that our boundary samples produce different labels on different architectures makes them interesting new tools for adversarial ML in general.

%


\bibliographystyle{IEEEbib}
\bibliography{references}

\end{document}